# Dialysis Risk Prediction and Treatment Effect Estimation for AKI patients using Longitudinal Electronic Health Records

Kalyani P. Pande, MS, Evan Yang, BS, Bryan Zhu, Sandeep K. Mallipattu, MD, Alisa Yurovsky, PhD, Tengfei Ma, PhD
Stony Brook University, Stony Brook, NY, USA

**Abstract**
*Progression to dialysis or end-stage renal disease is a rare but clinically important outcome. Clinicians need evidence on how medication exposures influence downstream risk. We constructed a fixed-window EHR cohort (90-day observation, 730-day prediction; N=81401; dialysis/ESRD prevalence: 1.1%) and modeled sequences of diagnoses, procedures, and medications with kidney laboratory trends (creatinine, BUN, eGFR). A transformer-based causal multi-head model was trained to estimate drug- and ingredient-level average treatment effects (ATEs) using counterfactual exposure removal and insertion under a full medication history setup. On test set, predictive performance reached an AUC of 0.694 and PR-AUC of 0.094. At the selected decision threshold (0.883), the model achieved an F1 score of 0.201 with a Brier score of 0.018. Post-hoc causal analyses of lab changes (eGFR, creatinine, BUN) using IPTW, AIPW, naive, and covariate-adjusted OLS methods assessed clinical directionality. Results showed partial protective-direction support for ACE/ARB exposures and worsening-direction signals for loop diuretics.*

**Introduction**
Progression to dialysis or ESRD is a major public health challenge associated with increased morbidity, mortality, and healthcare use. Long-term studies show that patients experiencing acute kidney injury (AKI) face significantly higher risks of progressing to chronic kidney disease, requiring dialysis, or death compared to those without AKI[1]. Many AKI patients undergo rapid clinical changes and need early detection to support timely referral, medication review, and monitoring. Clinicians also need stronger evidence on how medications affect kidney outcomes. While electronic health record (EHR) data provide detailed records of diagnoses, lab results, and medication exposures over time, identifying medication effects from observational data remains difficult. Traditional predictive models often identify correlations rather than causation and can be influenced by confounding factors and fragmented exposure signals[2].

In nephrology practice, treatment decisions are usually based on active ingredients and pharmacologic mechanisms rather than specific drug products. However, assessing medication effects is challenging because of polypharmacy, which is common among patients with kidney disease and other related conditions[3]. Patients often receive multiple therapies at once, making it hard to determine whether medications are linked to disease severity or directly cause changes in kidney outcomes.

In this work, we treat dialysis risk prediction and medication effect estimation as a causal inference problem. We analyze how predicted risk shifts under hypothetical exposure scenarios, combining sequence modeling with lab trends to capture changing physiological states. A transformer-based architecture simultaneously learns treatment assignment and outcome prediction. Ingredient-level average treatment effects (ATEs) are calculated by modifying medication exposure tokens and comparing the predicted risks across different hypothetical scenarios. To help address confounding, we compare treatment-effect signals with observed kidney function changes, allowing us to contrast model-based effects with physiological indicators of kidney injury.

Contributions
This study makes several methodological and empirical contributions to the analysis of kidney outcomes using electronic health record (EHR) data.
1. We develop a reproducible longitudinal cohort construction pipeline to predict 2-year dialysis risk for AKI patients with 90 days or 365 days observation window. This design supports consistent temporal ordering of exposures and outcomes while enabling time-aware evaluation.
2. We develop a unified framework for both dialysis risk prediction and treatment effect estimation based on transformers. We introduce a transformer-based causal modeling framework that jointly learns dialysis risk and treatment assignment from longitudinal EHR histories. The proposed multi-head architecture captures sequential clinical context while enabling the model to represent both outcome prediction and treatment-selection patterns within a unified representation.

3. We estimate medication effects at the active-ingredient level using counterfactual intervention strategies. By modifying medication exposure tokens during inference, we estimate ingredient-level average treatment effects (ATEs) expressed as differences in predicted dialysis risk under alternative exposure scenarios. This approach provides interpretable summaries of medication effects while reducing fragmentation caused by product-level medication coding.

4. We incorporate evaluation strategies tailored to rare clinical outcomes. Model performance is assessed using time-aware validation procedures and imbalance-aware training approaches, alongside causal baseline comparisons and overlap diagnostics to support interpretation of treatment-effect estimates.

5. We introduce an additional clinical validation layer based on kidney laboratory markers. Adjusted time-anchored analyses of changes in estimated glomerular filtration rate (eGFR), creatinine, and blood urea nitrogen (BUN) are used to evaluate whether inferred medication effects align with clinically consistent directions observed in nephrology practice.

Related Work

A growing body of research has applied structured electronic health record and claims data to predict adverse kidney outcomes, including acute kidney injury, chronic kidney disease progression, and dialysis initiation. These studies highlight several methodological challenges. First, dialysis and ESRD outcomes are relatively rare events, which can lead to severe class imbalance where high overall accuracy may coexist with poor recall of minority cases. Second, temporal leakage can occur when models incorporate information recorded after early manifestations of clinical deterioration, thereby inflating apparent performance. To mitigate these issues, prior work has emphasized the use of chronological train–test splits, fixed-window cohort designs, and clearly defined lookback periods that more closely reflect real-world clinical deployment[4-7]. Most previous studies focused on predicting kidney failure risk for CKD progression but rarely investigated prediction models specifically for AKI patients[1].

Sequence modeling approaches have also been widely applied to EHR data. Early models such as RETAIN demonstrated that attention-based recurrent neural networks can achieve strong predictive performance while providing interpretable representations of clinical histories[8]. More recent work has adopted transformer-based architectures, including BEHRT and Med-BERT, which capture long-range dependencies and complex temporal patterns in large-scale EHR data[9,10]. These approaches build upon the self-attention mechanism introduced in the transformer architecture[11].

Estimating treatment effects from observational health data has a long history in causal inference research. Classical approaches emphasize the use of propensity scores, regression adjustment, and sensitivity analyses to address confounding in observational studies[12,13]. More recently, representation-learning approaches for counterfactual inference have been proposed, including neural architectures that learn shared patient representations alongside treatment-specific outcome predictors[14,15]. Despite these methodological advances, treatment-effect estimation from EHR data remains challenging due to treatment-selection bias, measurement error, and time-varying confounding.

Medication interactions and clinical contexts also play an important role in kidney outcomes. Prior computational research has explored modeling drug–drug interactions using graph-based representations and network models of polypharmacy effects[16].

Evaluation of models for rare clinical outcomes requires careful consideration of appropriate performance metrics. For low-prevalence events such as dialysis initiation, precision–recall behavior, recall, and calibration are often more informative than accuracy alone[17,18]. Approaches such as focal loss and class-balanced objectives have been proposed to improve model learning under severe class imbalance[19,20]. In addition, calibration diagnostics such as Brier scores are commonly used to evaluate whether predicted probabilities align with observed outcome frequencies[21].

**Methods**

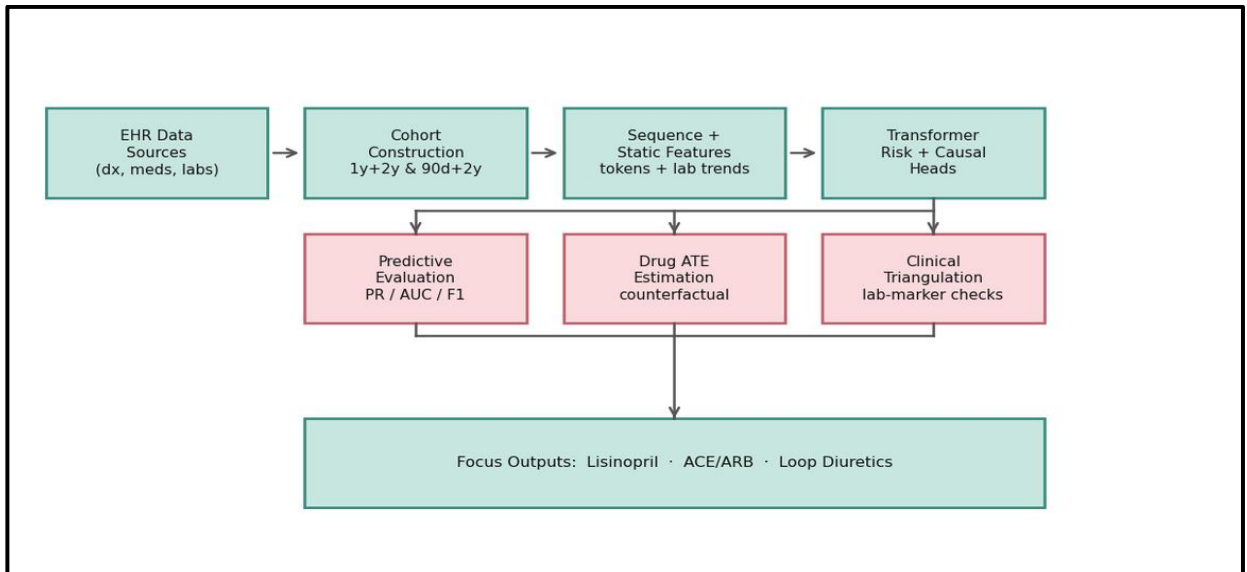

**Figure 1.** Pipeline for dialysis risk prediction and treatment effect estimation from EHR data.

Dataset and Cohort Construction
We utilized a de-identified electronic health record (EHR) dataset from the TriNetX public research network, which compiles longitudinal clinical data from participating healthcare providers. The dataset includes structured tables on patient encounters, diagnoses, procedures, medication records (covering both drug products and ingredient-level details), and lab results. A dialysis-risk group was created using a fixed-window approach, with a 90-day observation period followed by a 730-day prediction window. Patients had to meet minimum follow-up duration and encounter count criteria to ensure accurate representation of their clinical history. Additionally, a 365-day window was included to evaluate performance metrics.

Patients were labeled as outcome-positive ($Y = 1$) if dialysis or ESRD events occurred strictly after the observation period and within the prediction window. Events during the observation period were excluded to prevent label leakage and maintain a clear temporal distinction between predictors and outcomes. Each patient was represented as a sequence of diagnosis, procedure, and medication codes, with relative day offsets from the cohort index date. These sequences retain the chronological order of clinical events within the observation window and are used as the main input for subsequent sequence modeling and treatment-effect analysis.

Representation and Preprocessing
Separate vocabularies were constructed for diagnosis, procedure, and medication codes using the training dataset. Event sequences were ordered by relative time within the observation window and padded within minibatches to enable efficient model training. Medication exposures were represented using the full medication history observed within the observation window. In addition to event sequences, we incorporated summary features derived from laboratory measurements to capture physiological trajectories related to kidney function.

Specifically, laboratory trend features were derived from eGFR, creatinine, blood urea nitrogen (BUN), and estimated glomerular filtration rate (eGFR). For each laboratory marker, we calculated the last observed value, mean value, and slope across the observation window. These laboratory-derived features were combined with simple healthcare utilization summaries, including counts of diagnosis, procedure, and medication events, to provide low-dimensional contextual information representing baseline clinical status.

Model Architecture: Transformer-Based Causal Model
We developed a transformer-based causal modeling architecture to represent longitudinal EHR histories. Diagnosis, procedure, and medication events were embedded using separate token vocabularies, along with segment or type embeddings to differentiate event categories. Positional embeddings encoded the temporal order within sequences. The embedded event sequences were concatenated and processed through a shared transformer encoder with stacked self-attention layers. The resulting patient representation was pooled and fed into multiple prediction heads. The model includes four output heads: an outcome head predicting dialysis risk, a propensity head modeling treatment assignment, and two treatment-specific outcome heads corresponding to potential outcomes under treated and untreated conditions. This multi-head design allows joint modeling of outcome prediction and treatment-selection patterns within a shared representation.

The model was implemented in PyTorch. The transformer encoder consisted of four layers with a hidden dimension of 128 and four attention heads, with dropout regularization during training. Optimization was performed using the Adam optimizer with minibatch stochastic gradient descent. Hyperparameters were chosen based on validation experiments on the training split.

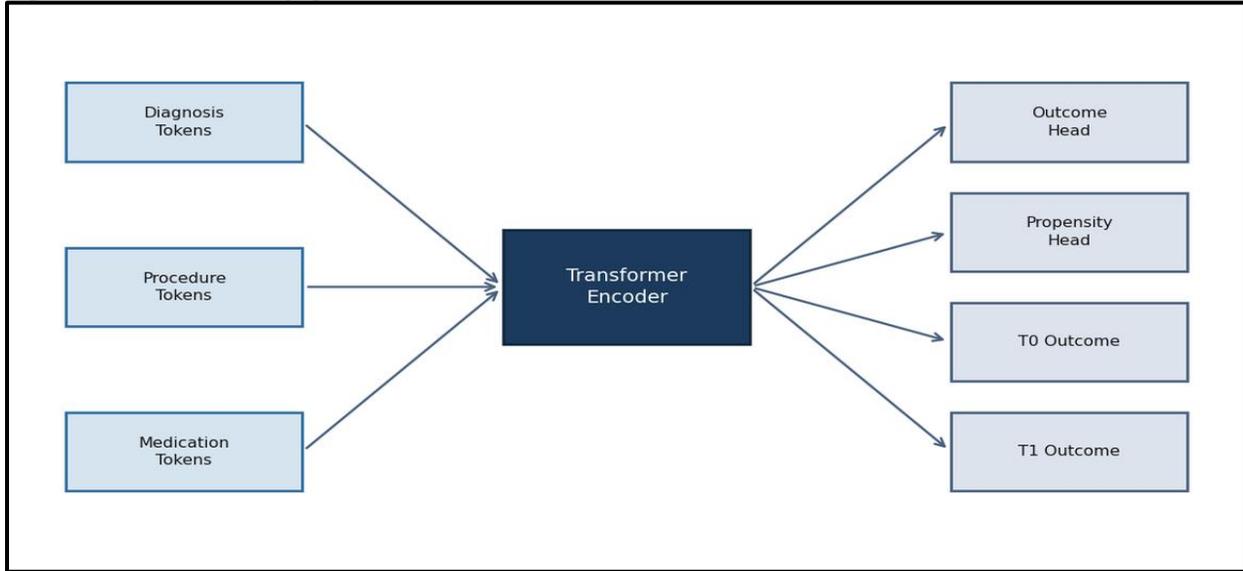

**Figure 2.** Transformer architecture for dialysis risk prediction and counterfactual Treatment effect estimation.

Training Objectives and Class Imbalance Handling
The model supports two training settings. In the primary setting, the model is trained to predict dialysis risk using the outcome head. In an optional treatment-specific setting, the model is jointly trained to model treatment assignment and potential outcomes. For the primary dialysis risk prediction task, the outcome head is optimized using an imbalance-aware binary classification objective. Depending on the configuration, this objective may include weighted binary cross-entropy, focal loss, or class-balanced loss functions designed to address rare-event classification.

When treatment assignment is explicitly modeled, the propensity head and potential outcome heads are trained jointly with the outcome prediction objective. The factual prediction is defined as $\hat{y}_t = t\hat{y}_1 + (1-t)\hat{y}_0$. The model is optimized using a combined loss function $L = L_y(\hat{y}_t, y) + \alpha L_t(\hat{t}, t) + \lambda \|\theta\|_2^2$ where $L_y$ represents the outcome loss, $L_t$ represents the propensity loss, α controls the contribution of the treatment assignment objective, and $\lambda$ represents L2 regularization.

To mitigate extreme outcome imbalance, we employ imbalance-aware loss functions and optionally apply weighted sampling to increase representation of positive cases during minibatch training. Limited token-level augmentation techniques, including dropout, masking, and sequence perturbation, are applied primarily to positive examples to improve model robustness.

Validation Protocol
Model evaluation was conducted using time-aware train, validation, and test splits to prevent temporal leakage. Predictive performance metrics were calculated on the held-out test cohort. For ingredient-level treatment-effect analyses, we optionally performed nested K-fold evaluations on the held-out dataset to assess the stability of estimated treatment effects across folds. These analyses summarize variability in estimated effects using descriptive statistics, including mean, standard deviation, and range.

Predictive performance metrics include the area under the receiver operating characteristic curve (AUC), precision–recall metrics, and the Brier score to evaluate probability calibration. Operating thresholds were chosen using the validation set by identifying the threshold on the precision–recall curve that maximized the F1 score, with optional constraints prioritizing recall.

Treatment Effect Estimation

To estimate medication effects, we used a counterfactual inference framework that involved modifying medication exposures within patient sequences. For each target drug or ingredient, predicted dialysis risk was calculated under two hypothetical exposure scenarios: one representing treatment and the other representing no treatment. The average treatment effect (ATE) was defined as the average difference in predicted risk between these two scenarios across patients. Treatment exposure was identified as the presence of a target ingredient in the patient's medication history during the observation period.

Counterfactual scenarios were created by changing medication exposure tokens within the patient's sequence while keeping all other historical data the same. In the untreated scenario, tokens for the target medication were removed or replaced with padding. In the treated scenario, the target medication token was added to the sequence if it was not already there. Ingredient-level effects were estimated by associating each ingredient with its set of medication codes and applying the same counterfactual modifications at the group level. This method allows for estimating treatment effects at both the drug-product and active-ingredient levels, enabling comparisons between detailed medication representations and clinically meaningful ingredient groupings. Effect estimates were reported alongside exposure support, which refers to the number of patients meeting the treatment criteria, and included additional stability diagnostics when applicable.

Post-hoc Time-Anchored Kidney-Marker Validation

To contextualize model-derived treatment-effect estimates, we conducted a post-hoc validation based on changes in kidney laboratory markers between baseline and follow-up. Baseline values were the latest lab measurements during the observation window, while follow-up values were the first measurements during the prediction window.

Changes in eGFR, creatinine, and BUN assessed whether medication signals aligned with expected kidney function changes. For each medication, treatment exposure was defined as its presence during the observation window. Various statistical estimators, including unadjusted comparisons, propensity-score weighting, doubly robust estimators, and covariate-adjusted regression, estimated associations between treatment and lab changes. Covariates included baseline labs, healthcare utilization, and medication burden proxies. Propensity score distributions and overlap diagnostics assessed comparison validity. Confidence intervals used bootstrap resampling. P-values were adjusted with Benjamini–Hochberg for multiple comparisons. Consistent directional changes across kidney markers provided stronger evidence for medication effects.

**Results**

**Table 1.** Cohort characteristics and laboratory availability.

| Characteristic | Value |
| --- | --- |
| N patients | 81401 |
| Dialysis or ESRD outcome prevalence | 1.10% |
| Age, mean (SD) | 51.9 (16.0) |
| Observation window | 90 days |

| Prediction window | 730 days |
|---|---|
| Creatinine available | 4.10% |
| BUN available | 39.40% |
| eGFR available | 93.5% |

Cohort Characteristics and Outcome Prevalence

The final cohort included 81,401 patients, with an observed dialysis or ESRD outcome prevalence of 1.10%. The low outcome prevalence highlights the substantial class imbalance inherent in dialysis prediction and motivates the use of imbalance-aware training strategies and recall-oriented evaluation metrics.

The mean patient age was 51.9 years (SD 16.0). Each patient contributed a 90-day observation window followed by a 730-day prediction window. Laboratory availability varied across kidney markers. Creatinine measurements were available for 4.10% of patients, eGFR measurements were available for 93.5%, while BUN measurements were available for 39.40%. These availability patterns reflect the heterogeneity of routine clinical data capture and limit the extent of laboratory-based validation analyses.

Predictive Performance

**Table 2.** Predictive performance of the transformer model on the test set for two different observation windows.

| Metric | Transformer | | Linear SVM | | Logistic Regression | |
|---|---|---|---|---|---|---|
|  | **90 days observation** | **365 days observation** | **90 days observation** | **365 days observation** | **90 days observation** | **365 days observation** |
| AUC | 0.694 | 0.657 | 0.495 | 0.573 | 0.665 | 0.697 |
| PR-AUC | 0.094 | 0.079 | 0.014 | 0.037 | 0.045 | 0.054 |
| Precision | 0.209 | 0.159 | 0.023 | 0.088 | 0.099 | 0.120 |
| Recall | 0.193 | 0.221 | 0.084 | 0.119 | 0.144 | 0.159 |
| F1 Score | 0.201 | 0.185 | 0.036 | 0.101 | 0.118 | 0.137 |
| Decision Threshold | 0.883 | 0.207 | 0.577 | 0.185 | 0.767 | 0.762 |
| Brier Score | 0.019 | 0.026 | - | - | 0.138 | 0.260 |

Overall, our Transformer-based method outperforms two commonly used baselines on almost all metrics (with rare exceptions): SVM and Logistic Regression.

With 90-day observation window, it achieved an AUC of 0.694 and a precision–recall AUC of 0.094, indicating moderate discrimination in a highly imbalanced setting. At a threshold of 0.883, it reached a precision of 0.209, recall of 0.193, and an F1 score of 0.201. The Brier score of 0.019 suggests reasonable calibration for rare-event prediction. Interestingly, extending the observation window to 365 days makes the prediction results worse for the transformer model, probably because it introduced more irrelevant noisy medical codes.

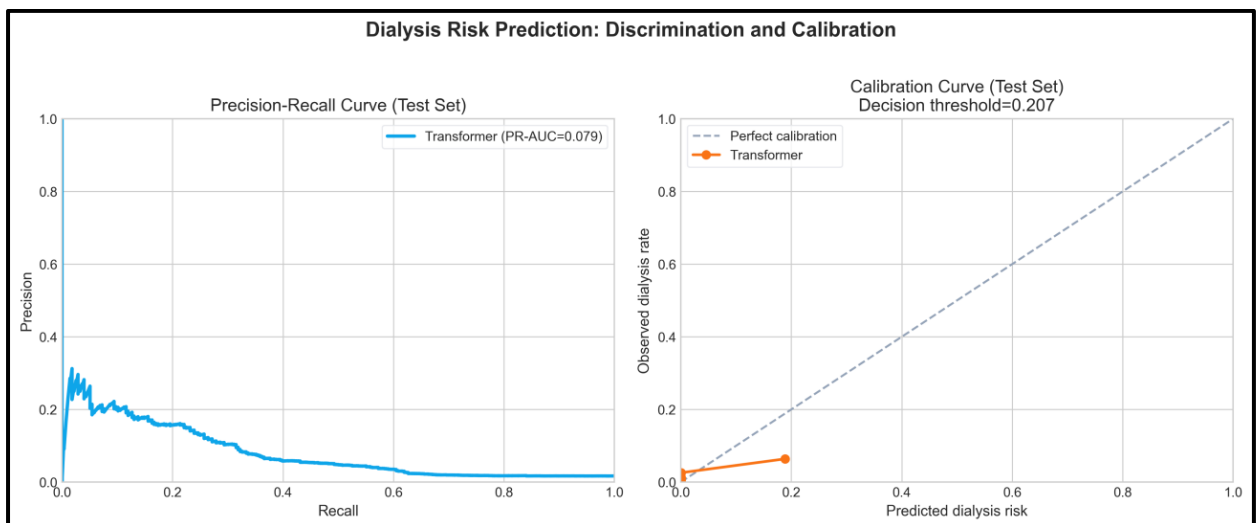

**Figure 3.** Precision-recall curve for dialysis risk prediction on the test set and calibration curve comparing predicted and observed dialysis risk.

Ablation and Sensitivity

**Table 3.** Ablation study evaluating architectural and training components.

| Variant | Recency decay | AUC | PR-AUC | F1 |
| --- | --- | --- | --- | --- |
| Base transformer | 0 | 0.58195853 | 0.02235463 | 0.04239213 |
| Lab trends | 0 | 0.52479231 | 0.02037808 | 0.03920118 |
| Imbalance handling | 180 | 0.58157392 | 0.04980705 | 0.11442786 |

We further conducted an ablation study to evaluate the relative contribution of architectural and training components within the modeling framework. The naive transformer model achieved moderate predictive performance under a simplified configuration. Incorporating imbalance-aware training strategies substantially improved precision–recall AUC and F1 score, highlighting the importance of explicitly addressing outcome rarity in dialysis prediction tasks. Additional variants explored the impact of laboratory trend features and recency-decay mechanisms. Although recency decay improved certain precision–recall metrics relative to simpler variants, the final model retained full longitudinal history without recency decay, as this configuration provided a more consistent representation of patient trajectories. Because these ablation experiments were conducted under a standardized lightweight training budget, their results are intended primarily to compare relative architectural contributions rather than absolute predictive performance.

Ingredient/Drug ATEs

We estimated medication effects using a counterfactual exposure framework in which predicted dialysis risk was compared under exposure insertion versus exposure removal scenarios. Resulting average treatment effects (ATEs) represent the predicted difference in dialysis risk attributable to a specific medication exposure. Ingredient-level ATE estimates are visualized in Figure 5, and selected results are summarized in Table 4.

Several medications within the ACE inhibitor / angiotensin receptor blocker (ACEi/ARB) class including lisinopril-related codes showed protective-direction effects, corresponding to lower predicted dialysis risk under exposure scenarios.

In contrast, loop diuretic medications demonstrated risk-increasing directional effects, with positive ATE values suggesting higher predicted dialysis risk in exposure scenarios. These findings are consistent with clinical expectations regarding the underlying disease severity of patients receiving these therapies.

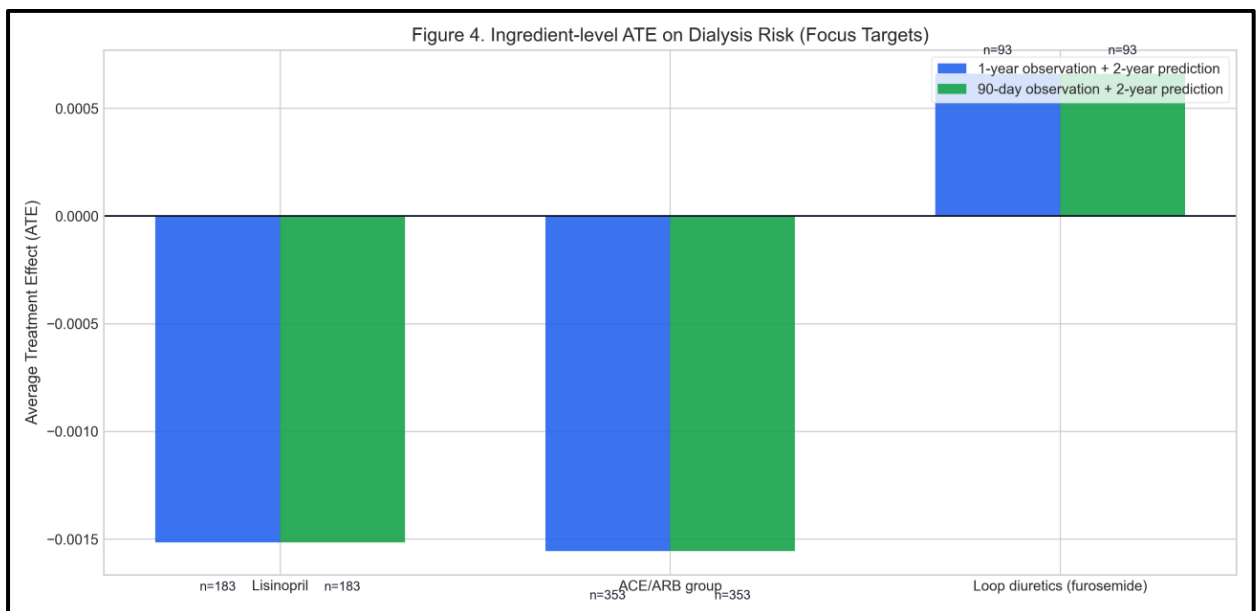

**Figure 5.** Ingredient-level Average Treatment effects (ATEs) on dialysis risk.

Given the potential for confounding by indication, these model-derived ATE estimates are interpreted as hypothesis-generating signals rather than definitive causal estimates. To further examine the clinical plausibility of these signals, we conducted additional validation using time-anchored kidney marker analyses.

**Table 4.** Estimated Ingredient-level treatment effects on Dialysis risk for Two different observation windows.

| Drug/ Ingredient | ATE | Direction | Support | Consistency |
|---|---|---|---|---|
| Lisinopril | -0.002 | Protective | 183 | Partial |
| ACE/ ARB group | -0.002 | Protective | 353 | Mixed |
| Loop Diuretics | +0.007 | Risk increasing | 210 | Mixed |

Time-Anchored Kidney-Marker Validation
To evaluate whether model-inferred medication effects were directionally consistent with known renal physiology, we conducted adjusted time-anchored analyses of kidney function markers, including creatinine, eGFR, and BUN.

1. ACEi/ARB-focused protective validation (lisinopril-centered)
Angiotensin-converting enzyme inhibitors and angiotensin receptor blockers modulate the renin–angiotensin–aldosterone system, reducing intraglomerular pressure through efferent arteriole vasodilation. Clinical guidelines recommend temporary discontinuation during acute kidney injury but continuation during recovery because of their long-term renoprotective benefits[4,5]. Consistent with prior clinical literature, ACEi/ARB exposures in our dataset were associated with protective-direction signals in model-derived ATE estimates. When available, laboratory markers such as eGFR, creatinine and BUN generally exhibited stable or favorable directional patterns, though support was incomplete due to limited laboratory coverage. Overall, these observations provide partial empirical support for the protective interpretation of ACEi/ARB exposures.

2. Loop diuretics like furosemide increase urine output by inhibiting sodium transport in the loop of Henle and are commonly used for fluid overload in kidney injury. However, guidelines warn that excessive use may worsen renal function[4,5]. Our dataset shows loop diuretic exposure linked to increased risk signals and patterns indicating renal decline. These findings should be cautious, as such use often reflects disease severity rather than directly causing harm.

3. Several nephrotoxic drugs, including NSAIDs, aminoglycosides, amphotericin B, and antivirals, were rarely used in this dataset, likely due to clinical caution in kidney injury cases[4,5]. Limited exposure data prevent reliable effect estimates, so no definitive conclusions about their impact can be made.

Robustness and Sensitivity
To assess the stability of treatment-effect estimates, we performed nested-fold ATE summarization, subgroup analyses, and comparisons with multiple causal baseline estimators, including naïve adjustment, inverse probability weighting (IPTW), augmented inverse probability weighting (AIPW), targeted maximum likelihood estimation (TMLE), and doubly robust ATT-style estimators. For laboratory-based validation analyses, we also examined propensity score overlap diagnostics. Observations outside the common support region were flagged as potential overlap violations, and such cases were treated as caution signals when interpreting treatment effects. Treatment-effect estimation was restricted to medications represented in a curated medication catalog, ensuring consistent mapping between drug products and their active ingredients while preserving surrounding clinical context.

Category-Level Summaries
Aggregating results across clinically meaningful medication categories provides an additional high-level validation check. The clearest patterns were observed for two nephrology-relevant categories:
1. ACE/ARB medications (lisinopril-focused) demonstrated protective-direction signals, with the strongest effects observed at specific lisinopril code-level rows. Evidence from kidney markers was partial and heterogeneous across the broader ACE/ARB group.
2. Loop diuretics (furosemide) demonstrated consistent worsening-direction signals across available kidney markers.

In contrast, several well-known nephrotoxic medication classes were too sparsely represented in the dataset to support reliable validation analyses.

**Conclusion**
This study introduces a reproducible, medication- focused pipeline for estimating treatment effects on dialysis risk across AKI and CKD trajectories using longitudinal electronic health record (EHR) sequences. In a cohort of 81,401 patients with a dialysis or ESRD prevalence of 1.1%, the transformer-based framework achieved moderate predictive discrimination (AUC = 0.694; PR-AUC = 0.094; F1 = 0.201) while enabling counterfactual treatment- effect estimation at both the drug-product and active-ingredient levels. The modeling approach combines transformer-based sequence representation, imbalance-aware training strategies, and catalog-constrained treatment-effect scoring, leveraging the full longitudinal medication history to ensure consistent representation across both risk prediction and treatment- effect estimation tasks. Clinical triangulation using kidney function markers provided some directional validation for several medication classes. Specifically, ACE inhibitor and angiotensin receptor blocker exposures showed protective signals in model- derived treatment effects, with lisinopril serving as the most consistent example at the code level. Conversely, loop diuretics exhibited risk-increasing signals along with worsening trends in kidney markers. These results align broadly with existing nephrology literature but are subject to notable limitations, including confounding by indication, limited laboratory data, and overlap constraints in observational data. Overall, the proposed framework shows how transformer-based modeling of longitudinal EHR data can support hypothesis generation related to medication safety and kidney outcomes, while maintaining interpretability through ingredient-level aggregation. However, the findings should be viewed as exploratory rather than definitive causal evidence.

Future work will focus on external validation across additional health systems, improved causal inference methods for time-varying confounders, enriched modeling of laboratory trajectories and medication attributes (including dosage and route), and engineering enhancements to improve reproducibility and experimental efficiency. Collectively, these efforts aim to advance scalable approaches for studying medication effects and renal outcomes within large observational EHR datasets.